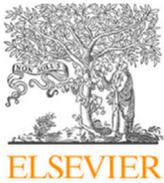
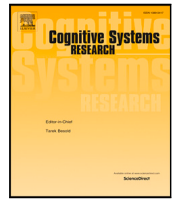

# Deep Robot Sketching: An application of Deep Q-Learning Networks for human-like sketching

Raul Fernandez-Fernandez [a,*], Juan G. Victores [b], Carlos Balaguer [b]

[a] ISCAR, Department of Computer Architecture and Automatics, Universidad Complutense de Madrid (UCM), Pl. de las Ciencias, 1, 28040 Madrid, Madrid, Spain
[b] Robotics Lab, Department of Systems Engineering and Automation, Universidad Carlos III de Madrid (UC3M), Avda. Universidad, 30, 28911 Leganés, Madrid, Spain

ABSTRACT

The current success of Reinforcement Learning algorithms for its performance in complex environments has inspired many recent theoretical approaches to cognitive science. Artistic environments are studied within the cognitive science community as rich, natural, multi-sensory, multi-cultural environments. In this work, we propose the introduction of Reinforcement Learning for improving the control of artistic robot applications. Deep Q-learning Neural Networks (DQN) is one of the most successful algorithms for the implementation of Reinforcement Learning in robotics. DQN methods generate complex control policies for the execution of complex robot applications in a wide set of environments. Current art painting robot applications use simple control laws that limits the adaptability of the frameworks to a set of simple environments. In this work, the introduction of DQN within an art painting robot application is proposed. The goal is to study how the introduction of a complex control policy impacts the performance of a basic art painting robot application. The main expected contribution of this work is to serve as a first baseline for future works introducing DQN methods for complex art painting robot frameworks. Experiments consist of real world executions of human drawn sketches using the DQN generated policy and TEO, the humanoid robot. Results are compared in terms of similarity and obtained reward with respect to the reference inputs.

## 1. Introduction

Artistic environments are considered within cognitive studies as rich, natural, multi-sensory, multi-cultural environments (Tversky, Healey, & Kirsh, 2014). Developing robots able to generate or imitate pieces of art has been a challenge that has always attracted the interest of the scientific robotic community and the general public. One proof of this is the RobotArt[1] annual competition with a total prize pool of $ 100 000 in 2018.

One of the scientific areas that has had more success developing art painting robots is the area of image processing. Multiple applications have been proposed combining state of the art image processing techniques and robot frameworks. An example of this is Paul, the portrait drawing robot (Tresset & Leymarie, 2013). Paul works by using an image processing algorithm to transform photos of people into human-like sketches. In this framework, high frequency lines are first extracted from the image. Then, a shadowing step is introduced to transfer the shadows from the image to the drawing. More recent frameworks introduce state of the art machine learning methods. An example of this is the introduction of Neural Style Transfer in painting robot applications. Neural Style Transfer is a technique proposed by Gatys, Ecker, and Bethge (2016). Here, Gatys proposed the introduction of Neural Networks to extract the style and content features of different images to perform a Style Transfer step between them. The result is a framework able to generate new images with the style of a selected artist. Many successful art painting robot applications have been proposed introducing this idea within their frameworks[1]. Other applications have also been proposed with the introduction of different machine learning techniques. An example of this is the framework proposed by Helou et al. using Autoencoders (El Helou, Mandt, Krause, & Beardsley, 2019). Here, the authors proposed a mobile robotic platform for big surface painting. Autoencoders were introduced in the pre-preprocessing step to obtain the desired painting texture. All these applications have in common that they focus on the implementation of state of the art image processing techniques to improve the quality and artistic value of the images generated in the pre-processing steps.

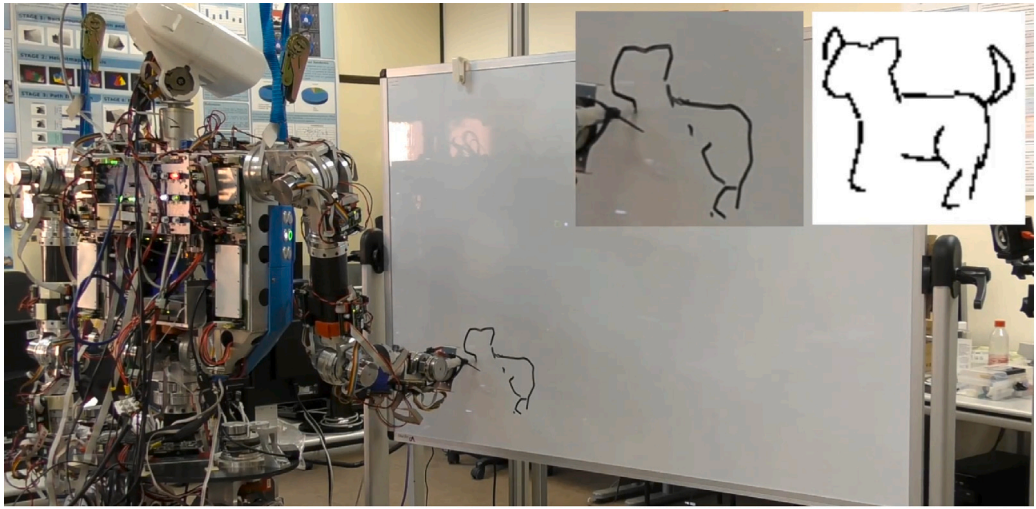

**Fig. 1.** Teo the humanoid robot drawing the sketch of a dog using the proposed framework.

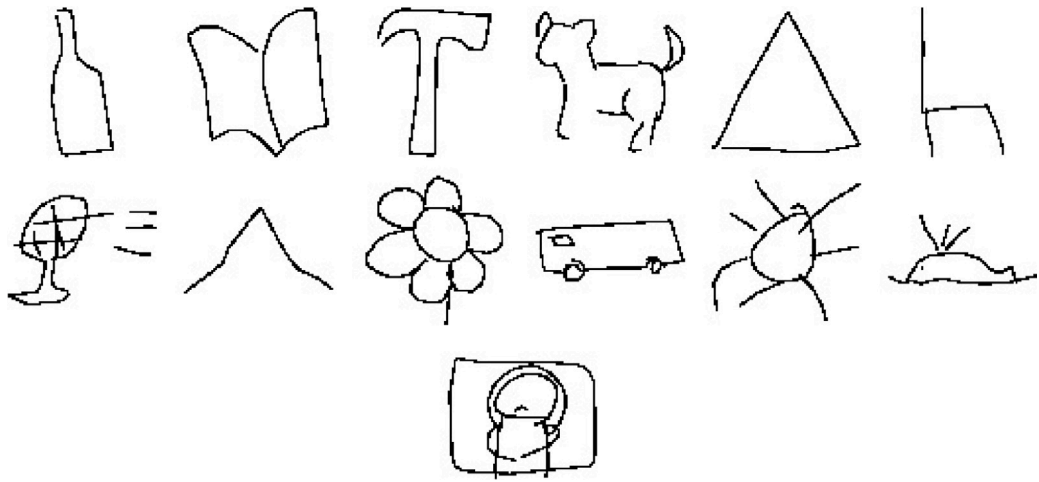

**Fig. 2.** Sketches extracted from the Quick, Draw! Dataset used for the experiments. From top left to bottom right: Wine bottle, book, hammer, dog, triangle, chair, fan, mountain, flower, bus, sun, whale and the Mona Lisa.

Reinforcement Learning is usually presented as an archetype of the success of recent theoretical approaches to cognitive science (Collins, 2019). Since the appearance of Deep Learning, Reinforcement Learning control policies are presented as an increasing popular way to implement robot control using Deep Neural Networks. Reinforcement Learning allows the autonomous learning of complex parametric control policies that can be implemented in complex robotics tasks (Haarnoja et al., 2019; Levine, Finn, Darrell, & Abbeel, 2016; Levine, Holly, Gu, & Lillicrap, 2017). Deep Q-learning Networks (DQN) are presented as one of the most successful architectures to implement Deep Reinforcement Learning in robotics (Levine et al., 2017). Zhou et al. (2018) successfully designed a DQN framework to generate sketches using the Quick, Draw! dataset (Jongejan, Rowley, Kawashima, Kim, & Fox-Gieg, 2016).

The introduction of a DQN algorithm within an art painting robot application using a humanoid robot is proposed in this paper. The goal is to study how DQN affects the performance of a basic art painting robot application. The main expected contribution of this work is to serve as a first baseline for future works introducing DQN methods for complex art painting robot frameworks.

The proposed DQN framework encodes the agent that defines the required pencil position within the canvas as a function of the current state. These generated agent positions are then transferred to a robotic platform via an internal robot cartesian gridmap for performing real world executions. The DQN framework proposed is based on the work by Zhou et al. (2018) and adapted to work with a real robot platform. As an additional contribution, a supplementary DQN classification network is defined within the reward function to improve the training step. The robotic platform selected for the experiments is TEO the humanoid robot from Universidad Carlos III de Madrid depicted in Fig. 1. Results are evaluated in terms of similarity with respect the reference sketches and the obtained DQN reward. Different parameters are introduced to measure and compare the complexity of each category. All of the reference sketches presented in this paper are extracted from the Quick, Draw! dataset. These selected sketches are shown in Fig. 2.

The structure of the paper goes as follows: first, an introduction and study of the related works is presented in Section 2. Section 3 depicts an introduction to the theoretical concepts involved in this paper. The proposed framework is defined in Section 4. Section 5 defines the experiments for the paper. Finally, results are presented in Section 6 and conclusions can be found in Section 7.

## 2. Related works

The study of the relevant related works is divided into two subsections: art and robots and art and machine learning. The first includes works where the goal is to design art painting robot applications that are able to generate or imitate pieces of art. The second focus on the study of the state of the art of machine learning frameworks that are related to the future development of art painting robot applications and focus on defining the full problem emphasizing on the execution step. The goal of the proposed work is to close this gap by introducing one of these ideas, a complex control DQN policy, within an art robot application.

*2.1. Art and robots*

Recent works, like the ones presented to the RobotArt[1] competition, have increased the quality of the pieces of art generated using robotic platforms. One of the reasons behind these improved results is the introduction of state of the art algorithms like Style Transfer. Some of these applications, however, do not provide any technical insight about the framework implementation. Due to the technical nature of this paper, only the robot frameworks with available scientific literature will be studied.

Earlier works, such as the already introduced Paul the robot (Tresset & Leymarie, 2013), have had already achieved impressive results using simple image processing techniques. More recently, Luo and Liu (2018) proposed a robot painting framework to draw cartoon-like portraits using image processing techniques such as face detection, facial decomposition and contour detection. The Busker robot (Scalera, Seriani, Gasparetto, & Gallina, 2018) proposed by Scalera et al. uses artistic rendering algorithms such as random stroke generation to generate watercolour paintings of 2D input images. Another interesting application is the one proposed by Alhafnawi, Hauert, and O'Dowd (2020) where a swarm of robots is used as pixels to create a painting canvas. In Karimov et al. (2021), three different processing image algorithms are introduced to generate artistic images from regular photos. In Chen et al. (2022) iLQR is used to generate an achievable robot trajectory using recorded artists painting motions and a cable-driven parallel robot.

Other works involving painting robots with different goals than art generation, have recently achieved higher levels of complexity by the introduction of Deep Neural Networks. This is the case of the already cited mobile robotic platform proposed by El Helou et al. (2019), and the "calligraphist" robot developed by Kotani and Tellex (2019). Few works involving art painting robots have focused on improving the robot execution step. These works propose the introduction of trajectory generation techniques similar to motion primitives (Makkar, Atoofi, Hamker, & Nassour, 2018; Mohan et al., 2011), or improving the low level motor control architectures as proposed by Atoofi, Hamker, and Nassour (2018).

*2.2. Art and machine learning*

Different machine learning frameworks have been proposed with the goal to create or imitate pieces of art. Most of these frameworks have been later used as the basis of art painting robot applications. In this section, a study of the state of the art of these machine learning frameworks is presented. Creative Adversarial Networks (CAN) is a framework proposed by Elgammal, Liu, Elhoseiny, and Mazzone (2017) using Generative Adversarial Networks (GAN) to produce new artificial generated artistic images. A more recent work by Mellor et al. (2019) also proposed the introduction of GAN frameworks for the generation of pieces of art. Here, instead of an image, the output of the framework are the strokes needed to paint that image. The results were obtained using an oil painting simulator (Li, 2023). A different approach was the already introduced framework proposed by Zhou et al. (2018). Here, the authors introduced the use of a DQN architecture for imitating human drawn sketches. More recently, Guo et al. (2020) proposed the introduction of a Learning from Demonstration framework to imitate the actions of a human painter using a robotic platform.

## 3. Background: Deep Q-learning neural networks

Let the task be defined as a Markov Decision Process (MDP) following the tuple $M = \{S, A, \gamma, R\}$: $S$ can be defined as a discretized state space of possible end-effector robot positions; $A$ can be defined as a set of possible actions inside this state space; the parameter $\gamma$ is the discount factor introduced to deal with long term predicted rewards, and $R$ is the reward function designed for the problem. The goal is to find a policy $\pi^*(s)$ that maximizes the expected reward of a full episode of the given task. Q-learning is presented as an off-policy Reinforcement Learning method that searches for the optimal policy $\pi^*(s)$ by maximizing the value function $Q(S, A)$ as defined in the Bellman equation in Eq. (1):

$$Q(S_t, A_t) = Q(S_t, A_t) + \alpha[R_{t+1} + \gamma \max_a Q(S_{t+1}, a) - Q(S_t, A_t)] \quad (1)$$

where $\alpha$ is the step-size parameter and $a$ is an action defined within the set of possible actions $A$.

The policy $\pi^*(s)$ can then be defined as the greedy policy that takes the action with the highest expected reward defined by $Q(S_t, A_t)$ at each time step $t$.

In DQN, the function $Q(S, A)$ is defined as a Deep Neural Network. The output of the Deep Neural Network is the Q-value $Q(s_t, A_t)$ corresponding to the state $s_t$ passed as input.

One of the main problems of Q-learning is the overestimation problem. This problem comes from the bias introduced by recurrently choosing and exploiting the action with the maximum current value, which may or may not be the action with the maximum average value. Double Q-learning (Van Hasselt, Guez, & Silver, 2016) is presented as a way to deal with this. The idea of Double Q-learning is to introduce an additional Q-learning function for the same problem. Only one of the two functions is selected to be updated each step. This selection can be performed at random and can change between steps. The Q-Value function is updated as in Eq. (2):

$$Q(S_t, A_t) = Q(S_t, A_t) + \alpha[R_{t+1} + \gamma \max_a Q'(S_{t+1}, a) - Q(S_t, A_t)] \quad (2)$$

where $Q'(S_{t+1}, a)$ is the estimated value of the not selected network. The introduction of this second network is demonstrated to deal with the overestimation problem and improve the convergence of the Q-learning function.

## 4. Framework

In this paper, an implementation of a DQN architecture for sketch drawing using a humanoid robot is proposed. This framework is based on the work introduced by Zhou et al. (2018) and adapted to work with a real humanoid robot.

*4.1. Model*

Two different inputs are introduced to the proposed DQN architecture depicted at Fig. 3. These inputs are defined as the global stream and the local stream. Each of these inputs are processed by a different part of the DQN architecture. The part of the network that processes the global stream is referred as the global network, while the part of the network that processes the local stream is referred as the local network. The outputs of the global and local networks are concatenated and processed by a fully connected network. This network is referred as the output network in the paper, and outputs the desired pen position within the local patch.

The global stream consists of four $M \times M$ channels where $M$ is the discretized canvas size. The first channel contains the generated canvas. This channel encodes, with a greyscale representation, the current state of the canvas where the agent is painting. The second channel is the reference canvas. This channel encodes the reference sketch using also a

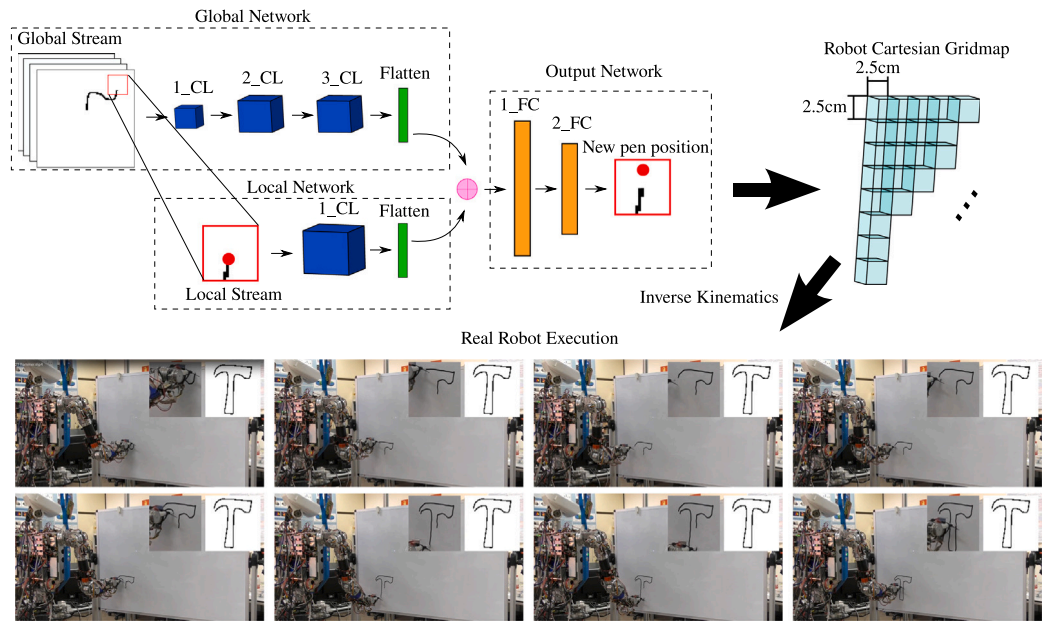

**Fig. 3.** Proposed DQN framework. The outputs of the global and local networks are concatenated and transferred to the output network. This output network generates the desired pen position within the local patch. This output is translated to a real robot cartesian gridmap. Using this robot cartesian gridmap, an Inverse Kinematic model is introduced to generate the control signals to move the robot to the position defined by the agent. TEO the humanoid robot from Universidad Carlos III de Madrid is introduced for the experimental setup.

**Table 1**
DQN layer specifications.

|  | Global network | | | Local network | Output network | |
| --- | --- | --- | --- | --- | --- | --- |
|  | 1_CL | 2_CL | 3_CL | 1_CL | 1_FC | 2_FC |
| Size | 32 | 64 | 64 | 128 | 512 | 242 |
| Kernel | 8 | 4 | 3 | 11 | – | – |
| Stride | 4 | 2 | 1 | 1 | – | – |

greyscale representation. The third channel is the distance map defined by Eq. (3):

$$D(x, y) = \frac{\sqrt{(x - p_x)^2 + (y - p_y)^2}}{M}, \forall (x, y) \in \Omega \quad (3)$$

where $\Omega$ depict the $M \times M$ canvas and $(p_x, p_y)$ is the current pen position within this canvas. This distance map contains the relative position of the pen with respect all the positions of the canvas. The last channel is the colour map. This colour map is fully set to ones (1) if the pen is touching the canvas or zeros (0) if it is not. Finally, the local stream is defined as a single $N \times N$ local patch of the generated canvas centred in the pen position, where $N \leq M$.

Different architectures are then introduced for the global and local network. The global network is composed of three convolutional layers (CL) with the following sizes: 32 with a kernel size of $8 \times 8$ and a stride of 4; 64 with a kernel size of $4 \times 4$ and a stride of 2; and 64 with a kernel size of $3 \times 3$ and a stride of 1. The local network is composed by a single convolutional layer with a size of 128 with a kernel size of $11 \times 11$ and a stride of 1. The output network takes as input the concatenated outputs of the global and local networks and is composed of two fully connected (FC) layers. The first layer has a size of 512 with a linear activation. The second and final layer has a size of 242 corresponding to all possible pen positions within the $11 \times 11$ patch for the painting and non painting states. The values of the different layers are depicted at Table 1.

*4.2. Training*

In order to improve the training process, a pre-training step is introduced to avoid training the DQN from scratch. In this pre-training step, artificial canvasses are generated using a random stroke generator. The network is trained using supervised learning with these artificially generated canvasses and the true ground actions required to generate them.

For the training step, Double Q-learning is implemented to avoid the overestimation problem. In terms of performance, an additional classification network with the same architecture as the one proposed in Fernandez-Fernandez, Victores, Estevez, and Balaguer (2019) is introduced to provide additional feedback to the agent. The resulting reward function using this classification network is defined in Eq. (4):

$$r = \begin{cases} r_k, & \text{if } k < pixel\_strokes \\ \Theta, & \text{otherwise if } k < total\_strokes \end{cases} \quad (4)$$

where the constant $pixel\_strokes$ is the number of strokes per sketch the agent is trained using $r_k$ and the $total\_strokes$ is the total number of strokes allowed to the agent per sketch, where $pixel\_strokes \geq total\_strokes$. The value $\Theta$ is the output of the classification network for the current sketch category. The reward $r_k$ at the $k_{th}$ step is defined as in Eq. (5)

$$r_k = r_{pixel} - P_{step} \quad (5)$$

where $P_{step}$ is a penalization value introduced to penalize slow agents. An agent is considered slow if, in one step, it moves less than 5 discretized positions in both dimensions. The $r_{pixel}$ value is defined as $r_{pixel} = s_k - s_{k+1}$ where $s_k$ is the similarity at the $k_{th}$ step of the generated canvas with respect the reference canvas as defined in Eq. (6):

$$s_k = \alpha * \frac{\sum_{i=1}^{M} \sum_{j=1}^{M} (P_{ij}^{ref} - P_{ij}^k)^2}{M^2} \quad (6)$$

where, the constant $\alpha$ is introduced to scale the similarity value with respect the output of the classification network. $P_{ij}^{ref}$ and $P_{ij}^k$ are the values at the $(i, j)$ position of the reference canvas and the generated canvas respectively.

*4.3. Integration with the real robot*

To transfer the position generated by the network for its execution in the real robot, a robot cartesian gridmap is introduced. This robot cartesian gridmap is the equivalent of the local patch, defined for the local stream, in the real world. This cartesian gridmap depicts the total 242 actions that can be executed by the agent. It is an $11 \times 11 \times 2$ (x, y, z) gridmap with 3D-cells of size $l_1 \times l_1 \times l_2$. The goal is to make the position of the agent within the local patch directly transferable to the real world. Each cell represents a possible cartesian robot end-effector position.

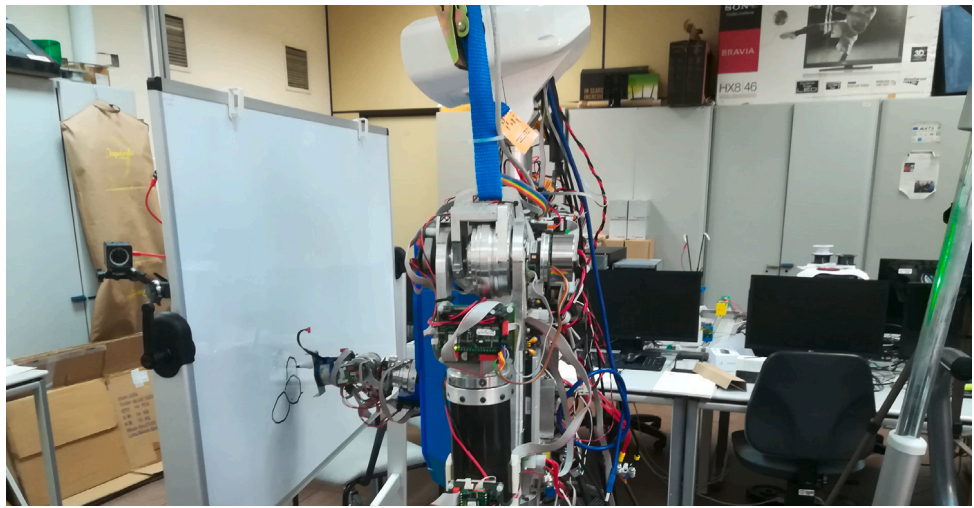

**Fig. 4.** Experimental setup used for the experiments. In the photo, TEO is drawing a flower.

In order to control the robot, for each new position selected by the agent ($p_{agent}$) within the local patch, a new cartesian robot end-effector position ($P_{cartesian}$) has to be obtained. If the robot cartesian gridmap is defined to be centred in the current robot end-effector position ($P_{current}$), the $P_{cartesian}$ position can then be obtained using Eq. (7):

$$P_{cartesian} = P_{current} + p_{agent} * l_1 \qquad (7)$$

This $P_{cartesian}$ position can then be used to control the robot via an Inverse Kinematic model. Changes in the colour selected by the agent (painting or not painting) are processed using the $z$ dimension. Non painting positions are defined with an $l_2$ offset in $z$ that moves the pen away from the canvas.

## 5. Experiments: Sketching with TEO

The humanoid robot TEO (Martínez et al., 2012) from Universidad Carlos III de Madrid was the selected robotic platform for the experiments. TEO is a 60 kg humanoid robot with 28 Degrees Of Freedom (DOF) and a height of 1.7 m. Each of the arms of the robot is composed by 6 DOF with a three-finger robot hand. For the experiments, the full right arm of the robot was used keeping the rest of the robot static. In order to allow the robot to paint, a marker was attached to the robot right hand. As the canvas, an erasable whiteboard was placed in front of the robot. The same robot starting position was used for all of the experiments. Fig. 4 depicts the experimental setup.

The sketches selected were part of the Quick, Draw! Dataset (Jongejan et al., 2016). Thirteen categories were randomly selected. These are, in no particular order: wine bottle, book, hammer, dog, triangle, chair, fan, mountain, flower, bus, sun, whale and the mona lisa. From these thirteen categories only eight were used for the training step: book, hammer, chair, fan, mountain, flower, bus and whale. The other five categories: wine bottle, dog, triangle, sun and the mona lisa were used only for testing. The training categories were divided in training and test datasets. The test categories were only used for testing. One random sketch for each of the categories was selected to be painted by TEO. These sketches were selected within the test dataset of each category. The reference selected sketches are shown in Fig. 2. The hyper-parameters used for the experiments are depicted at Table 2, where *max_total_strokes* is defined as the number of strokes used for training. The rest of the parameters are either defined within the definition of the algorithm or in the previous sections of this paper. These hyper-parameters are the result of multiple iterations of tuning the networks.

**Table 2**
Hyper-parameters for the experiments in this paper.

| Hyperparameter | Setting value |
|---|---|
| *Shared* | |
| Canvas size | $84 \times 84$ |
| Local patch size | $11 \times 11$ |
| Number of actions | 242 |
| Discount ($\gamma$) | 0.9 |
| Experience replay size | 10e3 |
| Batch size | 128 |
| Optimizer | Adam (Kingma & Ba, 2015) |
| *Pre-training* | |
| Epochs | 60e3 |
| Strokes per Epoch | Random number between 1–100 |
| Learning rate | 1e−5 |
| Loss function | Categorical Cross-Entropy |
| *total_strokes* | 100 |
| *Double Q-learning* | |
| Target update frequency | 10e3 |
| Epsilon | 0.1 |
| *pixel_strokes* | 100 |
| *total_strokes* | 150 |
| *max_total_strokes* | 15e4 |
| Learning rate | 1e−6 |
| Loss function | Mean squared error |
| Train dataset size | 3000 |
| Similarity scale | 1000 |
| *P_step* | 0.02 |

All the code has been open sourced and is available at https://github.com/roboticslab-uc3m/xgnitive-sketch.

## 6. Results

The experimental results obtained through the execution of the selected reference sketches using the proposed framework and TEO are depicted in Fig. 5 and Table 3. The waypoint agent trajectories depicted in the results are defined by the DQN agent positions. High level differences between reference sketches and executed sketches are introduced by the agent rather than by the robot. A video containing the execution of one of the sketches with the real robot is depicted in the following link https://youtu.be/YnSK9LMvIeU.

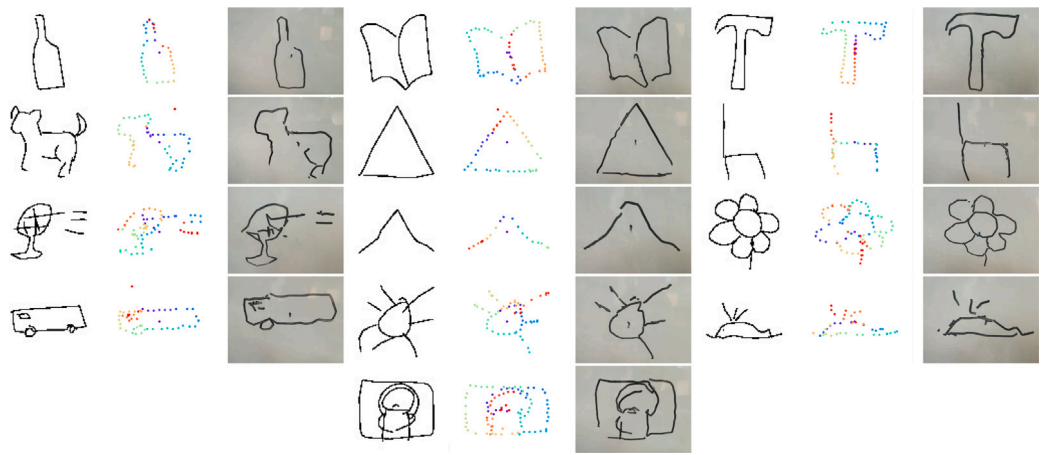

**Fig. 5.** Experimental results extracted from executing the DQN framework with TEO. Two resulting sketches are presented for each reference sketch. These are, from left to right, the waypoint trajectory defined by the agent and the robot executed sketch. The waypoint agent trajectory depicts the intermediate positions defined by the DQN agent. Purple colours depicts starting points, red colours depict the last points of the trajectory. The robot executed sketch is a photo of the resulting sketch drawn by the robot in the whiteboard.

**Table 3**
Experimental results.

| | MSE Simil. | DQN Rew. | Sketch strokes | Sketch points | Sketch Vel. | Categ. Strokes | Categ. Points | Categ. Vel. |
|---|---|---|---|---|---|---|---|---|
| Bottle | 92% | 22.06 | 1 | 25 | 30.92 | 2.48 | 25.50 | 34.32 |
| Book | 92% | 43.91 | 5 | 38 | 31.08 | 7.21 | 43.5 | 39.6 |
| Hammer | 93% | 37.88 | 1 | 31 | 31.26 | 3.33 | 34.19 | 36.36 |
| Dog | 88% | 38.79 | 7 | 54 | 23.76 | 7.18 | 61.42 | 23.72 |
| Triangle | 91% | 30.62 | 1 | 14 | 61.14 | 1.52 | 15.8 | 65.55 |
| Chair | 91% | 20.98 | 2 | 8 | 59.88 | 5.07 | 26.05 | 40.33 |
| Fan | 87% | 47.57 | 10 | 70 | 24.77 | 7.68 | 71.08 | 28.12 |
| Mountain | 92% | 17.05 | 1 | 7 | 98.14 | 2.1 | 19.05 | 50.49 |
| Flower | 89% | 61.29 | 4 | 65 | 19.6 | 4 | 61.63 | 23.61 |
| Bus | 88% | 31.88 | 4 | 38 | 28.32 | 7.36 | 60.15 | 25.78 |
| Sun | 89% | 35.90 | 8 | 37 | 25.32 | 8.1 | 42.68 | 26.76 |
| Whale | 89% | 26.87 | 6 | 55 | 23.96 | 7.36 | 60.15 | 25.78 |
| Mona Lisa | 84% | 68.5 | 8 | 79 | 36.76 | 8.79 | 64.82 | 32.33 |

Table 3 measures the similarity and DQN reward of the executed sketches. The similarity is obtained using a normalized Mean Squared Error (MSE) algorithm. The results depict high similarity values across all of the studied categories. The average similarity value in the case of the categories included in the training step is 90%. In the case of the categories introduced only for testing this value is 89%. The DQN Reward parameter depicts the total reward obtained by the agent during the full sketch execution. In order to compare results between different categories, three additional parameters were introduced to measure the complexity of each reference sketch and category. These three parameters are: the number of strokes and the number of points as defined in the Quick, Draw! dataset; and the stroke velocity (pixels/point). These parameters are obtained for each reference sketch and category. For the categories, these values are defined as the average value of all the sketches within the given category.

## 7. Conclusions

In this paper, an application of DQN within a robotic framework for drawing human-like sketches using the humanoid robot TEO has been proposed. The goal was to implement a Reinforcement Learning policy controller in a real art painting robotic framework. The robotic platform used was TEO, the humanoid robot from Universidad Carlos III de Madrid. The proposed framework takes as inputs multiple streams that encode the global and local features of the generated and reference canvasses. These inputs are defined as the global and local streams. Double Q-learning and an additional classification DQN are introduced for improving the training step. A real robot cartesian gridmap is defined as a way to transfer agent positions to real world cartesian coordinates. The framework was trained using eight categories extracted from the Quick, Draw! dataset. Five new categories were introduced only for testing. TEO, a multifunctional humanoid robot, a regular marker, and a conventional whiteboard were used as the only materials for the experiments.

Results show high similarity ratios for all of the studied categories. Relevant differences between the generated and the reference sketches only occur with the most complex sketches. Our hypothesis is that this happens due to the limitation in the number of strokes allowed by the agent. Complex sketches may require a higher number of maximum strokes. This maximum number of strokes was chosen to be the same as the one proposed in Zhou et al. (2018). In terms of the generalization capability of the network, the average similarity values obtained with the categories used only for testing are in the same range than the ones used also for training. Related future works involve the introduction of machine learning techniques to improve the creative step of the robot. This would allow the framework to generate its own artistic sketches and be executed in real time by the proposed DQN framework.

**Declaration of competing interest**

The authors declare that they have no known competing financial interests or personal relationships that could have appeared to influence the work reported in this paper.

**Data availability**

All the data in this paper has been open sourced and linked or referenced in the paper.

**Acknowledgments**

This research has been financed by ALMA, "Human Centric Algebraic Machine Learning", H2020 RIA under EU grant agreement 952091; ROBOASSET, "Sistemas robóticos inteligentes de diagnóstico y rehabilitación de terapias de miembro superior", PID2020-113508RB-I00, financed by AEI/10.13039/501100011033; "RoboCity2030-DIH-CM, Madrid Robotics Digital Innovation Hub", S2018/NMT-4331, financed by "Programas de Actividades I+D en la Comunidad de Madrid"; "iREHAB: AI-powered Robotic Personalized Rehabilitation", ISCIII-AES-2022/003041 financed by ISCIII and UE; and EU structural funds.

**References**

Alhafnawi, M., Hauert, S., & O'Dowd, P. (2020). Robotic canvas: Interactive painting onto robot swarms. In *ALIFE 2020: The 2020 conference on artificial life* (pp. 163–170). MIT Press, http://dx.doi.org/10.1162/isal_a_00285.

Atoofi, P., Hamker, F. H., & Nassour, J. (2018). Learning of central pattern generator coordination in robot drawing. *Frontiers in Neurorobotics*, http://dx.doi.org/10.3389/fnbot.2018.00044.

Chen, G., Baek, S., Florez, J.-D., Qian, W., Leigh, S.-W., Hutchinson, S., et al. (2022). GTGraffiti: Spray painting graffiti art from human painting motions with a cable driven parallel robot. In *2022 international conference on robotics and automation* ICRA, (pp. 4065–4072). IEEE, http://dx.doi.org/10.1109/ICRA46639.2022.9812008.

Collins, A. G. E. (2019). Reinforcement learning: bringing together computation and cognition. *Current Opinion in Behavioral Sciences*, *29*, 63–68. http://dx.doi.org/10.1016/j.cobeha.2019.04.011.

El Helou, M., Mandt, S., Krause, A., & Beardsley, P. (2019). Mobile robotic painting of texture. In *2019 international conference on robotics and automation* ICRA, (pp. 640–647). IEEE, http://dx.doi.org/10.1109/ICRA.2019.8793947.

Elgammal, A., Liu, B., Elhoseiny, M., & Mazzone, M. (2017). Can: Creative adversarial networks, generating "art" by learning about styles and deviating from style norms. https://arxiv.org/pdf/1706.07068.pdf.

Fernandez-Fernandez, R., Victores, J. G., Estevez, D., & Balaguer, C. (2019). Quick, stat!: A statistical analysis of the quick, draw! dataset. In *10th EUROSIM congress on modelling and simulation*. ARGESIM, http://dx.doi.org/10.11128/arep.58.

Gatys, L. A., Ecker, A. S., & Bethge, M. (2016). Image style transfer using convolutional neural networks. In *Proceedings of the IEEE conference on computer vision and pattern recognition* (pp. 2414–2423). IEEE, http://dx.doi.org/10.1109/CVPR.2016.265.

Guo, C., Bai, T., Lu, Y., Lin, Y., Xiong, G., Wang, X., et al. (2020). Skywork-daVinci: A novel CPSS-based painting support system. In *2020 IEEE 16th international conference on automation science and engineering* CASE, (pp. 673–678). IEEE, http://dx.doi.org/10.1109/CASE48305.2020.9216814.

Haarnoja, T., Ha, S., Zhou, A., Tan, J., Tucker, G., & Levine, S. (2019). Learning to walk via deep reinforcement learning. In *Robotics: Science and systems*. http://dx.doi.org/10.15607/RSS.2019.XV.011.

Jongejan, J., Rowley, H., Kawashima, T., Kim, J., & Fox-Gieg, N. (2016). *The quick, draw!-ai experiment, Vol. 17* (p. 4). Mount View, CA.

Karimov, A., Kopets, E., Kolev, G., Leonov, S., Scalera, L., & Butusov, D. (2021). Image preprocessing for artistic robotic painting. *Inventions*, *6*(1), http://dx.doi.org/10.3390/inventions6010019.

Kingma, D. P., & Ba, J. (2015). Adam: A method for stochastic optimization. In *3rd international conference on learning representations*. ICLR, https://arxiv.org/pdf/1412.6980.pdf.

Kotani, A., & Tellex, S. (2019). Teaching robots to draw. In *2019 international conference on robotics and automation* ICRA, (pp. 4797–4803). IEEE, http://dx.doi.org/10.1109/ICRA.2019.8793484.

Levine, S., Finn, C., Darrell, T., & Abbeel, P. (2016). End-to-end training of deep visuomotor policies. *Journal of Machine Learning Research*, *17*(1), 1334–1373. http://dx.doi.org/10.5555/2946645.2946684.

Levine, S., Holly, E., Gu, S., & Lillicrap, T. (2017). Deep reinforcement learning for robotic manipulation. U.S. Patent Application No. 16/333, 482.

Li, D. (2023). Fluid paint. http://david.li/paint/ (Accessed 17/01/2023).

Luo, R. C., & Liu, Y. J. (2018). Robot artist performs cartoon style facial portrait painting. In *2018 IEEE/RSJ international conference on intelligent robots and systems* IROS, (pp. 7683–7688). IEEE, http://dx.doi.org/10.1109/IROS.2018.8594147.

Makkar, D., Atoofi, P., Hamker, F., & Nassour, J. (2018). Motor program learning for humanoid robot drawing. In *2018 IEEE-RAS 18th international conference on humanoid robots (Humanoids)* (pp. 1–9). IEEE, http://dx.doi.org/10.1109/HUMANOIDS.2018.8624958.

Martínez, S., Monje, C. A., Jardón, A., Pierro, P., Balaguer, C., & Muñoz, D. (2012). Teo: Full-size humanoid robot design powered by a fuel cell system. *Cybernetics and Systems*, *43*(3), 163–180. http://dx.doi.org/10.1080/01969722.2012.659977.

Mellor, J. F., Park, E., Ganin, Y., Babuschkin, I., Kulkarni, T., Rosenbaum, D., et al. (2019). Unsupervised doodling and painting with improved spiral. https://arxiv.org/pdf/1910.01007.pdf.

Mohan, V., Morasso, P., Zenzeri, J., Metta, G., Chakravarthy, V. S., & Sandini, G. (2011). Teaching a humanoid robot to draw 'Shapes'. *Autonomous Robots*, *31*(1), 21–53. http://dx.doi.org/10.1007/s10514-011-9229-0.

Scalera, L., Seriani, S., Gasparetto, A., & Gallina, P. (2018). Busker robot: A robotic painting system for rendering images into watercolour artworks. In *IFToMM symposium on mechanism design for robotics* (pp. 1–8). Springer, http://dx.doi.org/10.1007/978-3-030-00365-4_1.

Tresset, P., & Leymarie, F. F. (2013). Portrait drawing by Paul the robot. *Computers & Graphics*, *37*(5), 348–363. http://dx.doi.org/10.1016/j.cag.2013.01.012.

Tversky, B., Healey, P. G. T., & Kirsh, D. (2014). Cognitive science and the arts. In *Proceedings of the 36th annual meeting of the cognitive science society, CogSci, Vol. 36* (pp. 58–59). https://cognitivesciencesociety.org/wp-content/uploads/2019/05/Program2014.pdf.

Van Hasselt, H., Guez, A., & Silver, D. (2016). Deep reinforcement learning with double q-learning. In *Proceedings of the AAAI conference on artificial intelligence, Vol. 30* (pp. 2094–2100). https://www.aaai.org/Press/Proceedings/aaai16.php.

Zhou, T., Fang, C., Wang, Z., Yang, J., Kim, B., Chen, Z., et al. (2018). Learning to doodle with stroke demonstrations and deep Q-networks. In *British machine vision conference 2018, BMVC*. BMVA, http://bmvc2018.org/contents/papers/0356.pdf.